\documentclass{article}
\usepackage{spconf,amsmath,graphicx}
\usepackage[utf8]{inputenc} 
\usepackage[T1]{fontenc}    
\usepackage{hyperref}       
\usepackage{url}            
\usepackage{booktabs}       
\usepackage{amsfonts}       
\usepackage{nicefrac}       
\usepackage{microtype}      
\usepackage{xcolor}         
\usepackage[numbers]{natbib}
\usepackage{amsmath}
\usepackage{svg} 
\usepackage{graphicx}
\usepackage{bm}
\usepackage{color}
\usepackage{tabularray}

\title{Spiking-Diffusion: Vector Quantized Discrete Diffusion Model with Spiking Neural Networks}
\name{Mingxuan Liu$^{\star}$
\qquad Jie Gan$^{\dagger}$
\qquad Rui Wen$^{\star}$ 
\qquad Tao Li$^{\star}$ 
\qquad Yongli Chen$^{\dagger}$ 
\qquad Hong Chen$^{\star}$ 
\thanks{Mingxuan Liu, Jie Gan and Rui Wen are co-first authors.}
\thanks{This work is supported by the National Science and Technology Major Project from Minister of Science and Technology, China (Grant No.
2018AAA0103100), and  the Laboratory Open Fund  of Beijing Smartchip Microelectronics Technology Co., Ltd (No. SGTYHT/21-JS-223).(Corresponding author: Hong Chen.)}}

 \address{$^{\star}$ Tsinghua University \\
      $^{\dagger}$ Beijing Smartchip Microelectronics Technology Co., Ltd}
\begin{document}
\topmargin=0mm
%
\maketitle
\begin{abstract}
Spiking neural networks (SNNs) have tremendous potential for energy-efficient neuromorphic chips due to their binary and event-driven architecture. SNNs have been primarily used in classification tasks, but limited exploration on image generation tasks. To fill the gap, we propose a Spiking-Diffusion model, which is based on the vector quantized discrete diffusion model. First, we develop a vector quantized variational autoencoder with SNNs (VQ-SVAE) to learn a discrete latent space for images. In VQ-SVAE, image features are encoded using both the spike firing rate and postsynaptic potential, and an adaptive spike generator is designed to restore embedding features in the form of spike trains. Next, we perform absorbing state diffusion in the discrete latent space and construct a spiking diffusion image decoder (SDID) with SNNs to denoise the image. Our work is the first to build the diffusion model entirely from SNN layers. Experimental results on MNIST, FMNIST, KMNIST, Letters, and Cifar10 demonstrate that Spiking-Diffusion outperforms the existing SNN-based generation model. We achieve FIDs of 37.50, 91.98, 59.23, 67.41, and 120.5 on the above datasets respectively, with reductions  of 58.60\%, 18.75\%, 64.51\%, 29.75\%, and 44.88\% in FIDs compared with the state-of-art work. Our code will be available at \url{https://github.com/Arktis2022/Spiking-Diffusion}.
\end{abstract}
\section{Introduction}
Spiking neural networks (SNNs) are the third generation neural networks with biological plausibility that encode and transmit information in form of spikes by mimicking the dynamics of neurons in the brain. Compared with Artificial neural networks (ANNs), the event-driven nature of SNNs allows for significant reduction in energy consumption when running on neuromorphic chips \cite{b6}.

SNNs with deep learning techniques have shown promising results on simple tasks such as image classification, image segmentation, and optical flow estimation\cite{snn}. However, the scope of their utility in complex tasks, particularly image generation, is still confined. Spiking-GAN \cite{b17} is the first fully SNN-based GAN that uses time-to-first-spike (TTFS) encoding to generate MNIST images through adversarial training. Despite its significant progress, the quality of the generated images still falls short of ANN-based GAN. Another model, fully spiking variational autoencoder (FSVAE) \cite{b22}, is a VAE model constructed entirely from SNN layers. This model produces comparable results to those of ANN-based VAEs, but its performance is still constrained by the inherent limitations of VAEs, namely the weak generative capacity due to the overly simplified decoder and compressed latent space \cite{b24}. Thus, more exploration and development are needed to harness the full potential of SNNs in image generation tasks. 

In this paper, we propose a vector quantized discrete diffusion model with spiking neural networks (called Spiking-Diffusion), which is the first diffusion model using only SNN layers. The pipeline of proposed Spiking-Diffusion contains two stages. First, an image is transformed into a discrete matrix through proposed vector quantized spiking variational autoencoder (VQ-SVAE). However, creating VQ-VAEs \cite{b3} in SNNs brings two challenges. The one is converting spike sequences into dense features suitable for storage in the codebook, because storing spike sequences directly within the codebook will consume too much memory. To overcome this challenge, we combine spike frequency rate (SFR) and postsynaptic potential (PSP) to model the spike sequences and introduce the learnable operator $k$ to optimize the weights of PSP and SFR. The other challenge is losslessly converting embedded features to spiking input for SNN decoder, because Poisson coding incurs information loss. To address this challenge, we design an adaptive spike generator (ASG) before the decoder and train it by using the same dictionary learning algorithm as traditional VQ-VAE. In the second stage, we utilize a spiking diffusion image decoder (SDID) to fit the prior discrete latent codes. In the forward process, we select a Markov transition matrix with an absorbing state \cite{b30} to gradually add masks to the discrete matrix, and adopt an SNN denoising network SDID to recover the masks and achieve inverse process parameterization. Experiments on MNIST \cite{b20}, FMNIST \cite{b31}, KMNIST \cite{b32}, Letters \cite{b33}, and Cifar10 \cite{cifar10} show that Spiking-Diffusion outperforms the SOTA SNN-based generative model.

\begin{figure*}[htb]
\centering
\includegraphics[width=0.8\textwidth]{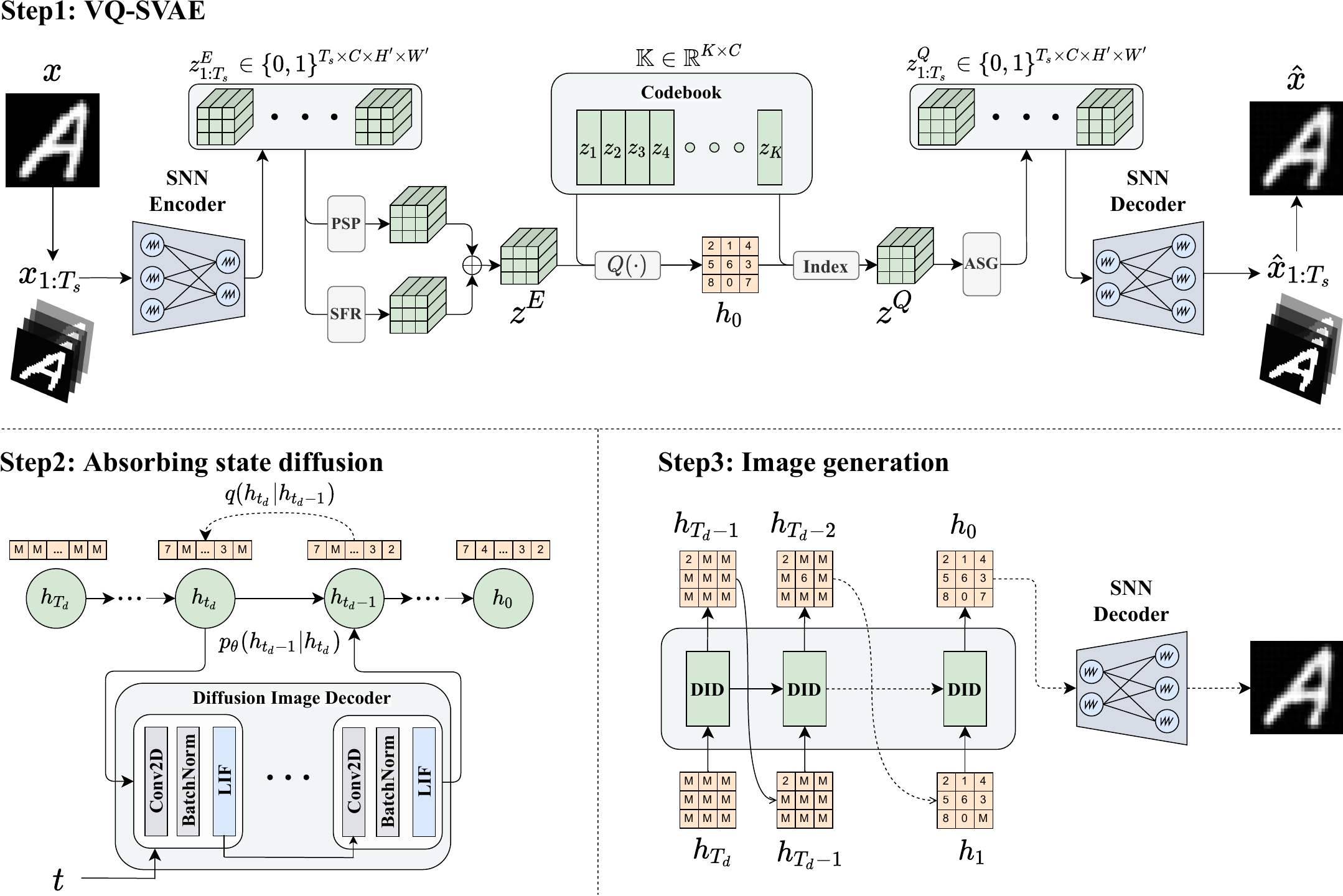}
\caption{The training process of Spiking-Diffusion consists of two stages. Step 1: Compress the images into discrete variables through VQ-SVAE. Step 2: The spiking diffusion image decoder (SDID) models the discrete latent space by reversing the forward diffusion process, which gradually adds masks in the discrete matrix through a fixed Markov chain. Step 3: During the test process, the SDID lifts the masks through an autoregressive process to obtain a discrete matrix with the target distribution.}
\label{fig1}
\end{figure*}

\section{Methodology}
\subsection{SNNs Learning Algorithms}
In this study, we adopt he SNNs learning algorithms in \cite{b41}. Spike neurons in SNNs emulates the behavior of biological neurons by generating discrete pulses, or spikes, in response to input stimuli. The LIF neuron model \cite{b42} is used, which is described by the following dynamic equation:
\begin{equation}
    H[t]=V[t-1]+\frac{1}{\tau}(X[t]-(V[t-1]-V_{reset}))
\end{equation}
\begin{equation}
    S[t]=\Theta(H[t]-V_{th})
\end{equation}
\begin{equation}
    V[t]=H[t](1-S[t])+V_{reset}S[t]
\end{equation}
where $\tau$ represents the membrane time constant, $X[t]$ denotes the synaptic input current at time step $t$, $H[t]$ represents the membrane potential of a neuron after charging but before firing a spike, and $S[t]$ denotes the spike generated by a LIF neuron when its membrane potential exceeds the discharge threshold $V_{th}$. $\Theta(v)$ is the Heaviside step function, which is equal to 1 when $v\geq0$ and 0 otherwise. $V[t]$ stands for the membrane potential after a spike event, which is reset to the $V_{reset}$ using a hard reset \cite{b43}. If no spike is generated, it is equal to $H[t]$.

\subsection{VQ-SVAE}
Although diffusion models \cite{b5} can generate high-quality images, they need to predict the noise added in the forward step. However, SNNs are difficult to fit and process continuous analog signals due to their discrete spike encoding \cite{b26}.  In order to solve the problem, we propose a VQ-SVAE to learn a discrete latent representation of images.

As shown in Fig. \ref{fig1}, Given a 2D image $x\in\mathbb{R}^{1 \times H\times W}$, we first convert it into sequences $x_{1:T_s}\in\mathbb{R}^{T_s \times 1 \times H\times W}$ using direct input encoding \cite{coding}, and then the $x_{1:T_s}$ is encoded as a spike sequence $z_{1:T_s}^E\in\ \{0,1\}^{T_s \times C \times H'\times W'}$ after the SNN encoder. In order to prevent the codebook from consuming too much memory, we utilize spike firing rate (SFR) and postsynaptic potential (PSP) to model the spike sequence. SFR has been proven in neurobiology to have the ability to represent information. For example, auditory nerves use SFR to encode steady-state vowels \cite{b53}.
\begin{equation}
{\rm{SFR}}(z_{1:T_s}^E)=\frac{1}{T_s}\sum^{T_s}_{t=1}z^E_t
\end{equation}
PSP simulates the response of postsynaptic neurons to the action potential (AP) sequence. Similarly, studies have shown that PSP is related to experience-dependent plasticity in the mammalian nervous system \cite{b54}. We use the following formula to update the PSP \cite{b55}: 
\begin{equation}
{\rm{PSP}}(z_{\leq t_s}^E)=(1-\frac{1}{\tau_{syn}})\times{\rm{PSP}}(z_{\leq t_{s-1}}^E)+\frac{1}{\tau_{syn}}\times z_t^E
\end{equation}
where $\tau_{syn}$ is the synaptic time constant and ${\rm{PSP}}(z_{\leq 0}^E)$ is set to 0. We use a trainable operator $k$ to allocate the weights of the PSP and SFR, so the features used for quantized encoding $z^E\in\ \mathbb{R}^{C \times H'\times W'}$ can be obtained from the following formula:
\begin{equation}
z^E=k\times{\rm{SFR}}(z_{1:T_s}^E)+(1-k)\times{\rm{PSP}}(z_{\leq T_s}^E)
\end{equation}
After obtaining $z^E$, we construct a codebook $\mathbb Z \in \mathbb R^{K \times C}$, with $z_k \in \mathbb Z$ and  $z_{i,j}^E \in z^E$. The quantized encoding of $z^E$, denoted as a discrete matrix $h_0$, can then be derived through nearest neighbor search. According to $h_0$, each spatial feature $z_{i,j}^E$ is mapped to its nearest codebook entry $z_k$ by indexing in $\mathbb Z$, thereby obtaining the embedding feature $z^Q$:
\begin{equation}
h_0=Q(z^E,\mathbb Z)={\rm{arg min}}_k||z^E_{i,j}-z_k||_2
\end{equation}
\begin{equation}
z_Q = {\rm{Index}}(\mathbb Z,h_0)
\end{equation}
However, the embedding feature $z^Q$ incorporates the SFR and PSP features of spike sequences. Consequently, it cannot be Poisson encoded \cite{b55} as input to the SNN decoder without loss of information. To solve the problem, we use a SNN layer to construct a trainable adaptive spike generator (ASG) that learns to restore $z^Q$ to the original spike sequence $z^Q_{1:T_s}$. At this stage, all parameters in the network can be trained end-to-end with the following loss function:
\begin{equation}
\begin{aligned}
&\mathcal L_{{\rm{VQ-SVAE}}}=||x-\hat{x}||_2+||z^E-z^Q||^2_2\\
+&(\sum^{T}_{t=1}||{\rm{PSP}}(sg[z_{\leq t_s}^E])-{\rm{PSP}}(z^Q_{\leq t_s})||^2\\
+&\beta\sum^{T}_{t=1}||{\rm{PSP}}(sg[z_{\leq t_s}^Q])-{\rm{PSP}}(z^E_{\leq t_s})||^2)
\end{aligned}
\end{equation}
In the formula, $sg[\cdot]$ denotes the stop gradient operation. Notably, compared with VQ-VAE \cite{b3}, VQ-SVAE imposes an additional constraint on training the ASG, which corresponds to the third term above. Moreover, the straight-through estimator propagates the gradients from the spike sequences $z_{1:T_s}^Q$ to $z_{1:T_s}^E$ during backpropagation, rather than propagating the gradients of the embedded features $z^Q$. Since spike sequences exhibit sparsity, directly computing the mean squared error (MSE) loss between sequences is not the optimal distance metric, therefore, we set the third term in Eq. (9) to maximum mean discrepancy (MMD), whose kernel function is PSP.
\subsection{Absorbing State Diffusion}
The training dataset is encoded into discrete matrices $h_0$ by the VQ-SVAE to train the spiking diffusion image decoder (SDID). For the forward diffusion process $q(h_{t_d}|h_{t_d-1})$, we use a transition matrix with an absorbing state \cite{b30} to convert all the elements in the discrete matrix into a mask after a fixed number of $T_d$ time steps. In detail, for each discrete random variable element $h_{t_d}(i,j)$ in $h_{t_d}$ with $K$ categories, denoting the one-hot version of $h$ with the row vector $\bm{h}$, we can define the forward process as: 
\begin{equation}
q(\bm{h}_{t_d}|\bm{h}_{t_d-1})={\rm {Cat}}(\bm{h}_{t_d};\bm{p}=\bm{h}_{t_d-1}\bm{Q}_{t_d})
\end{equation}
where ${\rm{Cat}}(\bm{h};\bm{p})$ is a categorical distribution parameterized by $\bm{p}$, and $[\bm{Q}_{t_d}]_{ij}=q(h_{t_d}=j|h_{t_d-1}=i)$ is the transition matrix of the forward Markov chain. We adopt absorbing state diffusion which is compatible with SNNs. Its transition matrix ${Q}_{t_d}\in \mathbb R^{(K+1) \times (K+1)}$  can be expressed as follows:
\begin{equation}
\begin{aligned}
{Q}_{t_d}=
\begin{bmatrix}
1-	\gamma_{t_d} & 0  & \cdots   & \gamma_{t_d}   \\
0 & 1-\gamma_{t_d}  & \cdots   & 	\gamma_{t_d}  \\
\vdots & \vdots  & \ddots   & \vdots  \\
0 & 0  & \cdots\  & 1  \\
\end{bmatrix}
\end{aligned}
\quad
\end{equation}
\begin{equation}
    \gamma_{t_d}=\frac{1}{T_d-t_d+1}
\end{equation}
The Eq. (11) indicates that at each timestep, each regular token has probability $\gamma_{t_d}$ of being replaced by the [MASK] token, and probability $1-\gamma_{t_d}$ of remaining unchanged, but the [MASK] token remains unchanged. In our work, we set the [MASK] token as $K$, as a result, $\bm{Q}_{t_d} \in \mathbb R ^{(K+1)\times (K+1)}$ and each token possesses $K+1$ states.  Therefore, the transition matrix with the absorbing state is non-zero only on the diagonal and the last column, so the sparsity of transition matrix will help to reduce the computational cost of the forward process.

The reverse process is defined as a Markov chain parameterized by $\theta$, which gradually denoises from the data distribution 
$p_{\theta}(\bm{h}_{0:T_d})=p(\bm{h}_{T_d})\prod_{t_d=1}^{T_{d}}p_{\theta}(\bm{h}_{t_d-1}|\bm{h}_{t_d})$ and is optimized by the evidence lower bound (ELBO). With $t^{{\rm{th}}}_{d}$ term the loss can be written as:
\begin{equation}
\mathcal{L}_{t_d}=D_{\rm{KL}}(q(\bm{h}_{t_d-1}|\bm{h}_0)||p_{\theta}(\bm{h}_{t_d-1}|\bm{h}_{t_d}))
\end{equation}

To reduce the randomness of training, SDID $S(h_{t_d}, t_d)$ is employed to predict $p_{\theta}(\bm{h}_{0}|\bm{h}_{t_d})$ rather than learn $p_{\theta}(\bm{h}_{t_d-1}|\bm{h}_{t_d})$ directly. As a result, variational bound reduces to:  
\begin{equation}
\mathbb E_{q(\bm{h}_{0})}\Bigg[\sum_{t_{d}=1}^{T_d}\frac{1}{t_{d}} \mathbb E_{q(\bm{h}_{t_d}|\bm{h}_0)}
\Big[\sum_{\bm{h}_{t_d}(i,j)=m}\log{p_{\theta}}(\bm{h}_0(i,j)|\bm{h}_{t_d})\Big]\Bigg]
\end{equation}

\begin{table}
\centering
\caption{Performance comparisons of Spiking-Diffusion(Ours) and FSVAE on SpikingJelly \cite{b41}.}
\label{tb1}
\begin{tblr}{
  cell{2}{1} = {r=2}{},
  cell{4}{1} = {r=2}{},
  cell{6}{1} = {r=2}{},
  cell{8}{1} = {r=2}{},
  cell{10}{1} = {r=2}{},
  cell{10}{3} = {},
  cell{10}{4} = {},
  cell{10}{5} = {},
  cell{10}{6} = {},
  cell{11}{3} = {},
  cell{11}{4} = {},
  cell{11}{5} = {},
  cell{11}{6} = {},
  hline{1-2,4,6,8,10,12} = {-}{},
}
Dataset & Model         & MSE $ \downarrow$ & SSIM $ \downarrow$ & KID $ \downarrow$ & FID $ \downarrow$ \\
MNIST   & FSVAE         & 0.023             & 0.219              & 0.054             & 90.57             \\
        & \textbf{Ours} & \textbf{0.006}    & \textbf{0.077}     & \textbf{0.018}    & \textbf{37.50}    \\
FMNIST  & FSVAE         & 0.023             & 0.378              & 0.070             & 113.2             \\
        & \textbf{Ours} & \textbf{0.011}    & \textbf{0.233}     & \textbf{0.055}    & \textbf{91.98}    \\
KMNIST  & FSVAE         & 0.054             & 0.452              & 0.272             & 166.9             \\
        & \textbf{Ours} & \textbf{0.014}    & \textbf{0.151}     & \textbf{0.068}    & \textbf{59.23}    \\
Letters & FSVAE         & 0.015             & 0.180              & 0.140             & 95.96             \\
        & \textbf{Ours} & \textbf{0.005}    & \textbf{0.078}     & \textbf{0.075}    & \textbf{67.41}    \\
Cifar10 & FSVAE         & 0.024             & 0.689              & 0.181             & 218.6             \\
        & \textbf{Ours}          & \textbf{0.009}    & \textbf{0.399}     & \textbf{0.121}    & \textbf{120.5}    
\end{tblr}
\end{table}
\begin{figure}[htb]
\centering
\includegraphics[width=0.45\textwidth]{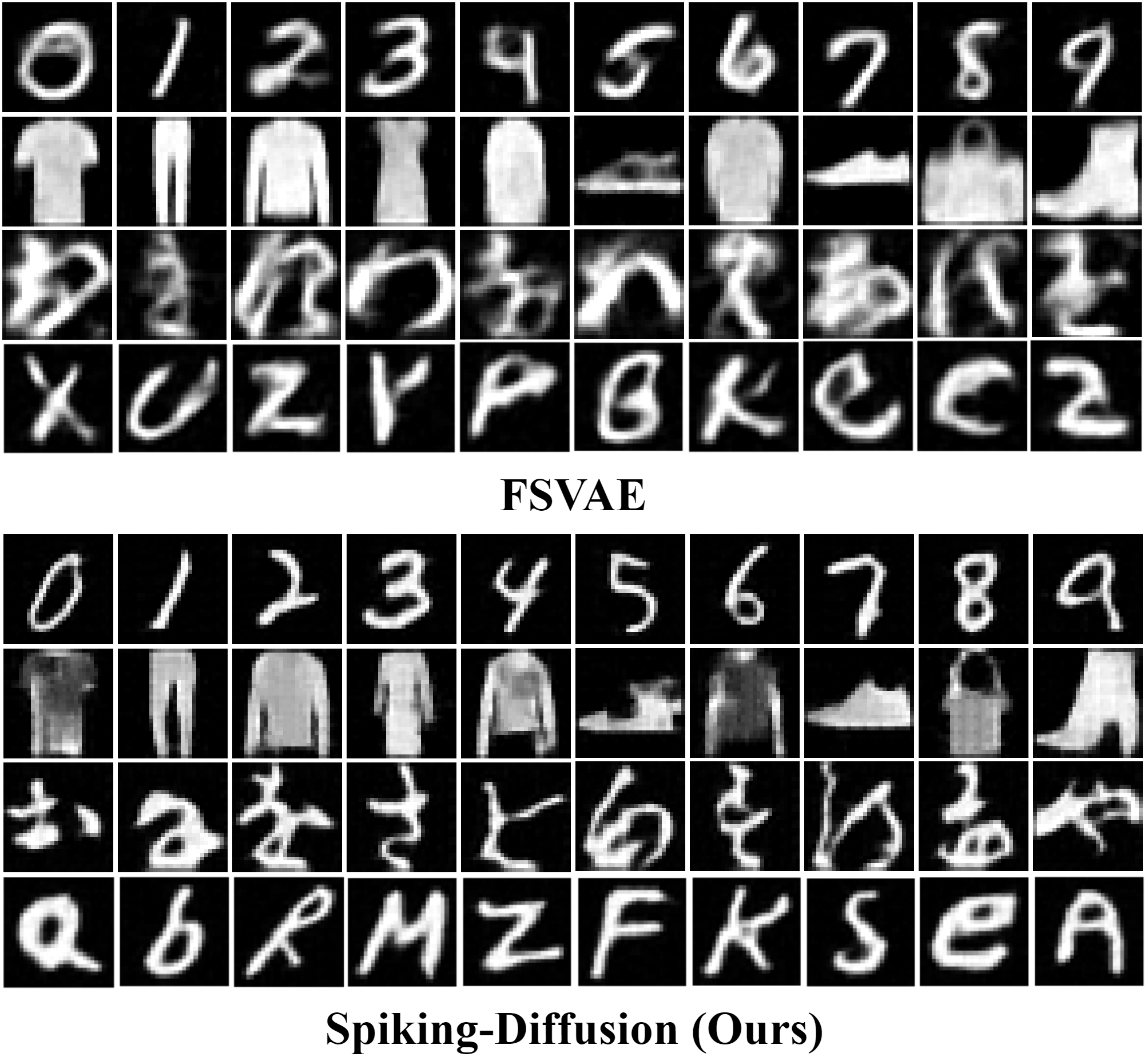}
\caption{Generated images of FSVAE and our Spiking-Diffusion (From top to down: MNIST, FMNIST, KMNIST, and Letters).}
\label{fig2}
\end{figure}

\section{Experiments}
\subsection{Datasets}
We test our model on five datasets: The MNIST \cite{b20} consists of 60,000 handwritten digit images divided into 10 classes (digits 0 to 9); FMNIST \cite{b31} contains 60,000 images of 10 different classes of clothing; KMNIST \cite{b32} consists of 60,000 images of Japanese characters, with each image belonging to one of 10 classes; Letters \cite{b33} provides 145,000 images of 26 Latin letters; for Cifar10\cite{cifar10}, 50,000 images are used for training and 10,000 images for evaluation.
\subsection{Implementation details}
We make a comprehensive comparison between Spiking-Diffusion and FSVAE \cite{b22}, which is the state-of-the-art SNN-based generative model generating images with quality equal to or better than ANN-based models. We test the FSVAE and Spiking-Diffusion on all datasets with the same experimental setup. The LIF neuron model's surrogate gradient function is $g(x)=\frac{1}{\pi}{\rm{arctan}}(\frac{1}{\pi}\alpha x)+\frac{1}{2}$, and its derivative is $g'(x)=\frac{\alpha}{2(1+(\frac{\pi}{2}\alpha x)^2)}$, where $\alpha$ represents the slope parameter. For all neurons, $\alpha=2$, $V_{reset}=0$, and $V_{th}=1$. The Optimizer is AdamW with $\beta=(0.9,0.999)$ and weight decay 0.001. The training lasts for 100 epochs. In addition, the original FSVAE used the STBP-tdBN \cite{b35} framework to train SNNs, while we adopt SpikingJelly \cite{b41} in our work. 

\subsection{Evaluation Metrics and Results}
\textbf{Evaluation Metrics.} We evaluate the reconstruction ability of the model by comparing the input and output images using two metrics: mean squared error (MSE) loss and structural similarity (SSIM) loss. To assess the quality of sampled images, we use two commonly used scores: Kernel Inception Distance (KID) and Fr\'{e}chet inception distance (FID). 

\textbf{Results.} Table \ref{tb1} shows that our Spiking-Diffusion model outperforms FSVAE in reconstruction and generation with the SpikingJelly framework. Spiking-Diffusion yields lower MSE and SSIM losses of reconstruction quality, and lower FID and KID scores of generated image quality. Figure \ref{fig2} shows the examples of generated images on the four datasets, from which we find that with the same SNN encoder and decoder structures, the images generated by FSVAE are blurrier, while the images generated by our Spiking-Diffusion are clearer with obvious boundaries. The is because that only the Bernoulli distribution is used in FSVAE to approximate the true posterior distribution, which cannot fully capture the rich semantic information in the image. In contrast, the Spiking-Diffusion can approximate the complex posterior distribution through nearest neighbor search and generate images by seeing context.

\section{Conclusion}
In this work, we present Spiking-Diffusion, the first implementation of diffusion models in SNNs. Our approach is based on VQ-DDM, which has two stages. First, we train the VQ-SVAE model to achieve image reconstruction, by using both PSP and SFR to model spiking features and constructing a codebook for image discretization. Secondly, we construct a discrete diffusion model in the discrete feature domain of images, and using absorbing diffusion state in Spiking-Diffusion is proven to generate highe quality images. The experimental results demonstrate that Spiking-Diffusion currently outperforms other SNN-based generative models. Future work will focus on how to train larger scale SNN generative models.

\bibliographystyle{IEEEbib}
\bibliography{strings,refs}

\begin{thebibliography}{10}

\bibitem{b6}
Jilin Zhang, Dexuan Huo, Jian Zhang, Chunqi Qian, Qi~Liu, Liyang Pan, Zhihua
  Wang, Ning Qiao, Kea-Tiong Tang, and Hong Chen,
\newblock ``22.6 anp-i: A 28nm 1.5 pj/sop asynchronous spiking neural network
  processor enabling sub-o. 1 $\mu$j/sample on-chip learning for edge-ai
  applications,''
\newblock in {\em 2023 IEEE International Solid-State Circuits Conference
  (ISSCC)}. IEEE, 2023, pp. 21--23.

\bibitem{snn}
João~D. Nunes, Marcelo Carvalho, Diogo Carneiro, and Jaime~S. Cardoso,
\newblock ``Spiking neural networks: A survey,''
\newblock {\em IEEE Access}, vol. 10, pp. 60738--60764, 2022.

\bibitem{b17}
Vineet Kotariya and Udayan Ganguly,
\newblock ``Spiking-gan: A spiking generative adversarial network using
  time-to-first-spike coding,''
\newblock in {\em 2022 International Joint Conference on Neural Networks
  (IJCNN)}. IEEE, 2022, pp. 1--7.

\bibitem{b22}
Hiromichi Kamata, Yusuke Mukuta, and Tatsuya Harada,
\newblock ``Fully spiking variational autoencoder,''
\newblock in {\em Proceedings of the AAAI Conference on Artificial
  Intelligence}, 2022, pp. 7059--7067.

\bibitem{b24}
Christos Louizos, Kevin Swersky, Yujia Li, Max Welling, and Richard Zemel,
\newblock ``The variational fair autoencoder,''
\newblock {\em arXiv preprint arXiv:1511.00830}, 2015.

\bibitem{b3}
Aaron Van Den~Oord, Oriol Vinyals, et~al.,
\newblock ``Neural discrete representation learning,''
\newblock {\em Advances in neural information processing systems}, vol. 30,
  2017.

\bibitem{b30}
Jacob Austin, Daniel~D Johnson, Jonathan Ho, Daniel Tarlow, and Rianne van~den
  Berg,
\newblock ``Structured denoising diffusion models in discrete state-spaces,''
\newblock {\em Advances in Neural Information Processing Systems}, vol. 34, pp.
  17981--17993, 2021.

\bibitem{b20}
Li~Deng,
\newblock ``The mnist database of handwritten digit images for machine learning
  research [best of the web],''
\newblock {\em IEEE signal processing magazine}, vol. 29, no. 6, pp. 141--142,
  2012.

\bibitem{b31}
Han Xiao, Kashif Rasul, and Roland Vollgraf,
\newblock ``Fashion-mnist: a novel image dataset for benchmarking machine
  learning algorithms,''
\newblock {\em arXiv preprint arXiv:1708.07747}, 2017.

\bibitem{b32}
Tarin Clanuwat, Mikel Bober-Irizar, Asanobu Kitamoto, Alex Lamb, Kazuaki
  Yamamoto, and David Ha,
\newblock ``Deep learning for classical japanese literature,''
\newblock {\em arXiv preprint arXiv:1812.01718}, 2018.

\bibitem{b33}
Gregory Cohen, Saeed Afshar, Jonathan Tapson, and Andre Van~Schaik,
\newblock ``Emnist: Extending mnist to handwritten letters,''
\newblock in {\em 2017 international joint conference on neural networks
  (IJCNN)}. IEEE, 2017, pp. 2921--2926.

\bibitem{cifar10}
Alex Krizhevsky, Geoffrey Hinton, et~al.,
\newblock ``Learning multiple layers of features from tiny images,''
\newblock 2009.

\bibitem{b41}
Wei Fang, Yanqi Chen, Jianhao Ding, Ding Chen, Zhaofei Yu, Huihui Zhou,
  Timothée Masquelier, Yonghong Tian, and other contributors,
\newblock ``Spikingjelly,''
  \url{https://github.com/fangwei123456/spikingjelly}, 2020,
\newblock Accessed: 2023-04-18.

\bibitem{b42}
RB~Stein and Alan~Lloyd Hodgkin,
\newblock ``The frequency of nerve action potentials generated by applied
  currents,''
\newblock {\em Proceedings of the Royal Society of London. Series B. Biological
  Sciences}, vol. 167, no. 1006, pp. 64--86, 1967.

\bibitem{b43}
Wei Fang, Zhaofei Yu, Yanqi Chen, Timoth{\'e}e Masquelier, Tiejun Huang, and
  Yonghong Tian,
\newblock ``Incorporating learnable membrane time constant to enhance learning
  of spiking neural networks,''
\newblock in {\em Proceedings of the IEEE/CVF International Conference on
  Computer Vision}, 2021, pp. 2661--2671.

\bibitem{b5}
Jonathan Ho, Ajay Jain, and Pieter Abbeel,
\newblock ``Denoising diffusion probabilistic models,''
\newblock {\em Advances in Neural Information Processing Systems}, vol. 33, pp.
  6840--6851, 2020.

\bibitem{b26}
Amirhossein Tavanaei, Masoud Ghodrati, Saeed~Reza Kheradpisheh, Timoth{\'e}e
  Masquelier, and Anthony Maida,
\newblock ``Deep learning in spiking neural networks,''
\newblock {\em Neural networks}, vol. 111, pp. 47--63, 2019.

\bibitem{coding}
Youngeun Kim, Hyoungseob Park, Abhishek Moitra, Abhiroop Bhattacharjee,
  Yeshwanth Venkatesha, and Priyadarshini Panda,
\newblock ``Rate coding or direct coding: Which one is better for accurate,
  robust, and energy-efficient spiking neural networks?,''
\newblock in {\em ICASSP 2022-2022 IEEE International Conference on Acoustics,
  Speech and Signal Processing (ICASSP)}. IEEE, 2022, pp. 71--75.

\bibitem{b53}
Murray~B Sachs and Eric~D Young,
\newblock ``Encoding of steady-state vowels in the auditory nerve:
  representation in terms of discharge rate,''
\newblock {\em The Journal of the Acoustical Society of America}, vol. 66, no.
  2, pp. 470--479, 1979.

\bibitem{b54}
Uma~R Karmarkar and Yang Dan,
\newblock ``Experience-dependent plasticity in adult visual cortex,''
\newblock {\em Neuron}, vol. 52, no. 4, pp. 577--585, 2006.

\bibitem{b55}
Friedemann Zenke and Surya Ganguli,
\newblock ``Superspike: Supervised learning in multilayer spiking neural
  networks,''
\newblock {\em Neural computation}, vol. 30, no. 6, pp. 1514--1541, 2018.

\bibitem{b35}
Hanle Zheng, Yujie Wu, Lei Deng, Yifan Hu, and Guoqi Li,
\newblock ``Going deeper with directly-trained larger spiking neural
  networks,''
\newblock in {\em Proceedings of the AAAI Conference on Artificial
  Intelligence}, 2021, pp. 11062--11070.

\end{thebibliography}

\end{document}